\documentclass{IEEE_lsens}
\pdfoutput=1
\usepackage{textcomp}

\usepackage[noadjust]{cite}

\ifCLASSINFOpdf
  
\else
  
\fi

\ifCLASSINFOpdf
\else
   \usepackage[dvips]{graphicx}
\fi
\usepackage{url}

\usepackage[dvipsnames]{xcolor}

\usepackage{graphicx}
\usepackage{amssymb}
\usepackage{amsmath}
\usepackage{times}
\usepackage{epsfig}
\usepackage{graphicx}
\usepackage{amsmath}
\usepackage{amssymb}
\usepackage{subcaption}
\newcommand{\ba}{\begin{array}}
\newcommand{\ea}{\end{array}}

\newcommand{\bc}{\begin{center}}
\newcommand{\ec}{\end{center}}
\newcommand{\bit}{\begin{itemize}}
\newcommand{\eit}{\end{itemize}}

\newcommand{\beq}{\begin{equation}}
\newcommand{\eeq}{\end{equation}}

\newcommand{\bfr}{\mathbf{r}}

\newcommand{\bfD}{\mathbf{D}}

\newcommand{\bfF}{\mathbf{F}}
\newcommand{\bfG}{\mathbf{G}}

\newcommand{\bfI}{\mathbf{I}}

\newcommand{\bfR}{\mathbf{R}}

\newcommand{\be}{\begin{equation}}
\newcommand{\ee}{\end{equation}}
\newcommand{\beqa}{\begin{eqnarray}}
\newcommand{\eeqa}{\end{eqnarray}}

\usepackage[T1]{fontenc} % optional enhanced font encoding
\usepackage{amsmath}
\usepackage[cmintegrals]{newtxmath}
\usepackage{bm}
\usepackage{array}
\usepackage{url}
\ifCLASSINFOpdf
\else
\fi
\providecommand{\hypersetup}[1]{\relax}

\hypersetup{pdftitle={Bare Demo of IEEE\_lsens.cls for IEEE Sensors Letters},%<!CHANGE!
pdfsubject={Typesetting},%<!CHANGE!
pdfauthor={Michael D. Shell},%<!CHANGE!
pdfkeywords={Class, IEEE, IEEE\_lsens, IEEE Sensors Letters, LaTeX, Typesetting, TeX}}%<^!CHANGE!}
\begin{document}

% \markboth{Vol.~1, No.~3, July~2017}{0000000}

% article subject line 
\IEEELSENSarticlesubject{IEEE Sensor Letters}

\title{RespVAD: Voice Activity Detection via Video-Extracted Respiration Patterns}

\author{\IEEEauthorblockN{Arnab Mondal\IEEEauthorrefmark{1} and Prathosh A.P\IEEEauthorrefmark{2}}% <-this % stops a space
\IEEEauthorblockA{\IEEEauthorrefmark{1,2}Department of Electrical Engineering,
Indian Institute of Technology, Delhi, New Delhi India.}%
% LSENS authors should provide a real e-mail address here.
%\thanks{Corresponding author: Arnab Kumar Mondal (e-mail: arnabkumarmondal123@gmail.com).\protect\\
}% <-this % stops a space
%
% note the % following lines that end in } - 
% these prevent an unwanted space from occurring between the objects.
% i.e., if you had this:
% For what it is worth, this is a minor point as most people would not even
% notice if the said evil space somehow managed to creep in.

% Manuscript received line
% \IEEELSENSmanuscriptreceived{Manuscript received June 7, 2017;
% revised June 21, 2017; accepted July 6, 2017.
% Date of publication July 12, 2017; date of current version July 12, 2017.}

\IEEEtitleabstractindextext{%
\begin{abstract}
Voice Activity Detection (VAD) refers to the task of identification of regions of human speech in digital signals such as audio and video. While VAD is a necessary first step in many speech processing systems, it poses challenges when there are high levels of ambient noise during the audio recording. To improve the performance of VAD in such conditions, several methods utilizing the visual information extracted from the region surrounding the mouth/lip region of the speakers' video recording have been proposed. Even though these provide advantages over audio-only methods, they depend on faithful extraction of lip/mouth regions. Motivated by these, a new paradigm for VAD based on the fact that respiration forms the primary source of energy for speech production is proposed. Specifically, an audio-independent VAD technique using the respiration pattern extracted from the speakers' video is developed. The Respiration Pattern is first extracted from the video focusing on the abdominal-thoracic region of a speaker using an optical flow based method.  Subsequently, voice activity is detected from the respiration pattern signal using neural sequence-to-sequence prediction models. The efficacy of the proposed method is demonstrated through experiments on a challenging dataset recorded in real acoustic environments and compared with four previous methods based on audio and visual cues. Data and the code for our implementation will be made available at https://github.com/arnabkmondal/RespVAD.
\end{abstract}

}

% \IEEEpubid{1949-307X \copyright\ 2017 IEEE. Personal use is permitted, but republication/redistribution requires IEEE permission.\\
% See \url{http://www.ieee.org/publications\_standards/publications/rights/index.html} for more information.}

\maketitle

\section{Introduction}

\IEEEPARstart{V}oice Activity Detection (VAD) is a critical first step in almost all speech processing applications \cite{ramirez2004new}. VAD is primarily a binary classification task where segments of input signals (such as audio and video) are categorized to be containing human speech activity or not \cite{sohn1999statistical}. This is a well studied problem and numerous algorithms have been proposed from time to time \cite{gerven1997comparative, lamel1981improved,drugman2015voice,van2013robust}. Recent techniques tackle the problem of VAD using deep learning models including Convolutional Neural Networks \cite{kaushik2018speech}, Recurrent Neural Networks \cite{gelly2017optimization}, WavNet-based architectures \cite{ariav2019end} and Encoder-Decoder networks \cite{ivry2019voice}.  
%\subsection{Prior Art: VAD from non-speech sources}
\par All the aforementioned methods utilize only the audio signal to perform the VAD. However, the process of human speech production results in multiple cues other than change in acoustic pressure. These include temporary distortion in the configuration of the vocal apparatus, apparent movement of the lips, change in the breathing patterns etc. Therefore it is imperative that some of these non-speech information be leveraged for performance enhancement of VAD. There have been several attempts towards using the video stream of a speaking subject to detect the presence of speech \cite{liu2004voice,sodoyer2006analysis,estellers2011dynamic,minotto2013audiovisual,dov2015audio,patrona2016visual,dov2016kernel}. These techniques first localize the lip region either through contour extraction \cite{sodoyer2006analysis,patrona2016visual} or by utilizing the distinctive color information \cite{lopes2011color,minotto2013audiovisual}. Subsequently, a set of  descriptors are extracted for the motion dynamics of the lip region \cite{dov2015audio,dov2016kernel}. These descriptors are then combined with the features extracted from the audio signals and used in conjunction with classifiers such as Gaussian Mixture models \cite{estellers2011dynamic,dov2015audio}, kernel machines \cite{dov2016kernel,patrona2016visual} and Deep neural networks \cite{sharma2019toward} for VAD. While all these methods are shown to enhance the performance of audio-only VADs, they are limited by the need to detect and localize the lip/mouth region as noted in \cite{sharma2019toward}. Motivated by these, a new paradigm for VAD based on the speech-induced changes in the breathing pattern is explored in this work. 

\begin{figure}
\centerline{\includegraphics[keepaspectratio,width=\columnwidth]{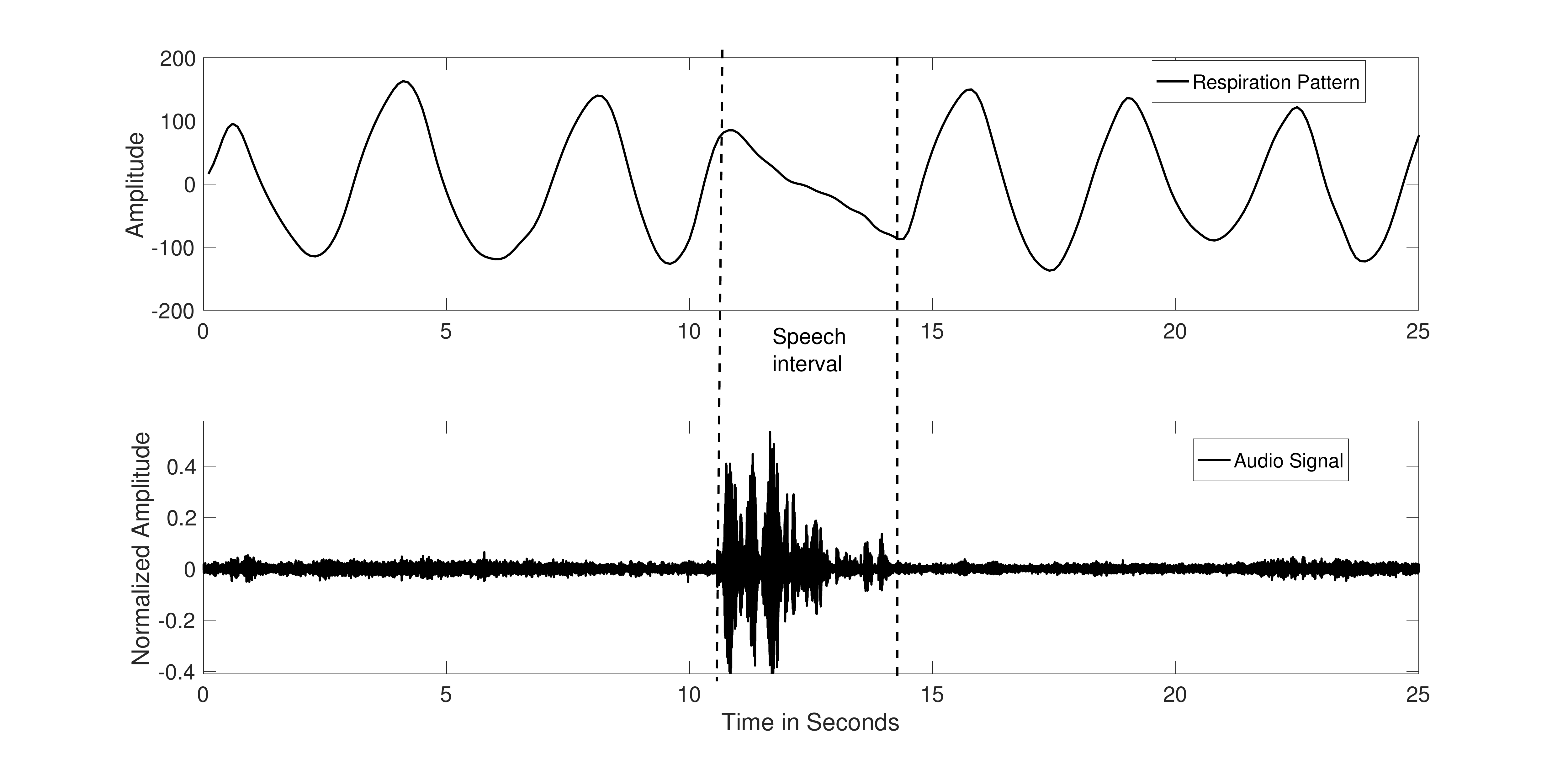}}
\caption{A segment of audio signal with the corresponding respiratory pattern. It is seen that the expiration cycle during presence of speech is distorted as compared to those during the non-speech phases.}
\label{speech_rp}

\end{figure}

\subsection{Role of Respiration in Speech Production}

Acoustic energy needed for speech production is primarily derived from the respiratory system \cite{draper1959respiratory,ohala1990respiratory}. The production of a typical speech utterance begins after a deeper inspiration, with the pressure of the air below the vocal cords being controlled by the descent of the rib cage. Subsequently, the lung volume decreases and reaches a stage where it is less than it is after a normal inspiration. At this instance, the relaxation pressure provides the energy required for a conversational utterance. After this, the expiration phase maintains the pressure below the vocal cords needed for sustaining the utterance \cite{ladefoged1988investigating}.  Fig. \ref{speech_rp} demonstrates a segment of audio signal with the corresponding respiratory pattern.  It is clearly seen that the expiration cycle during the presence of speech has a distinctive distortion as compared to those during the non-speech phases.  Given the aforementioned role of respiration in speech production, in this paper the following question is investigated: \textit{\textbf{Can the changes in the respiratory pattern during speech production be used for detection of the voice activity?}}
\par There have been multiple methods to extract the respiration pattern from a human subject including strain sensors \cite{chu2019respiration}, microphones \cite{lussu2020role} and video signals \cite{prathosh2017estimation}. However, in this work, video-based respiration extraction is considered since it is contact-free and requires no specialized hardware other than a video camera. Further, a large number of scenarios where VAD is necessary such as video conferencing, personal-assistants and cell-phone based speech recognizers, video recording is ubiquitous. This motivates the use of video-based respiration extraction albeit the VAD method described here is applicable on respiration patterns extracted from other modalities.

\begin{figure}[t]
    \centerline{\includegraphics[keepaspectratio,width=\columnwidth]{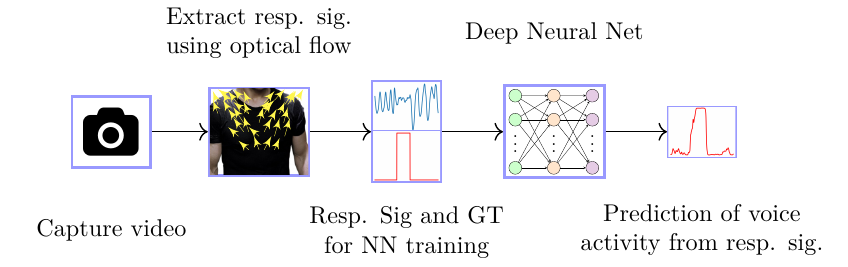}}
    \caption{Illustration of work flow of the proposed algorithm, RespVAD. The video of a speaking subject is captured from which the respiration pattern is extracted. A supervised  deep sequence-to-sequence model is trained to detect speech activity from respiration pattern.}
    \label{fig:bd}
\end{figure}
\section{Proposed Method}\label{sec:proposed_method}
\subsection{Extraction of Respiration Pattern from Video}
The first step in the proposed technique is to extract the Respiration Pattern (RP) signal. RP is extracted routinely in clinical settings using invasive medical devices such as impedance pneumography \cite{seppa2010assessment}.  However, such devices are infeasible in consumer applications such as VAD. Owing to this, an algorithm \cite{AC} where the RP signal is extracted from the video recording of a subject is proposed. The following is the description of the RP extraction algorithm: \\
A video recording of a human subject, roughly focusing on the abdominal-thoracic region, forms the input for the method. Suppose the video contains $N$ frames. Let $\bfI_t(x,y)$ denote the image intensity of the pixel of a frame at time $t$ and position $(x,y)$. It is assumed that the video scene has no other major source of motion than that caused by the respiration. With this, the RP signal is estimated as the one that is maximally correlated with motion (flow) vectors at all pixels. %\footnote{There is no need for a specific region of interest extraction as long as the video focuses roughly on the abdominal-thoracic region over which the assumption of absence of any other source of motion also becomes valid \cite{AC}.}.  
Let the temporal and the spatial gradients at every pixel $(x,y)$  of each frame be denoted as $\bfD_t(x,y)$ and $\bfG_t(x,y)$, respectively.
\be
\bfG_t(x,y)=\left[\begin{array}{c}\bfI_t(x+1,y)-\bfI_t(x,y)\\\bfI_t(x,y+1)-\bfI_t(x,y)\end{array}\right]
\ee
\be
\bfD_t(x,y)=\bfI_t(x,y)- \bfI_{t-1}(x,y)
\ee
The normalized directional optical flow, quantifying the change in the image gradient at every pixel, is defined as:
\be
\bfF_t(x,y)=\frac{\bfD_t(x,y) \bfG_t(x,y)}{\|\bfG_t(x,y)\|^2}
\ee
Let $ \bfF = [\bfF_1,...,\bfF_N] $ denote the matrix of flattened normalized flow vectors for all $N$ frames of the video. With this, $\bfR^* = [\bfr_1,...,\bfr_N]$, the $N$-length Respiratory Pattern that is to be extracted from the video, is given as the vector ($\bfR$) that is maximally correlated with all the rows of the $\bfF$ matrix: 

\begin{equation}
\bfR^* = \underset{{\bfR:\ ||\bfR||=1}}{\arg\max} \ \   ||\bfF\bfR||^2
\label{eqn:objectivefn}
\end{equation} where $||.||$ denotes any matrix norm. The RP  estimated this way maximizes its squared correlation with all the flow sequences corresponding to all pixels. It is easy to see that the solution for the optimization problem in Eq. 4 is given by (Refer to the supplementary material for derivation):
\begin{equation}
\bfR^* = \frac{1}{\sigma}\mathbf{U}_1^T\bfF
\label{eqn:optimalR}
\end{equation}
where $\mathbf{U}_1$ is the left singular vector corresponding to the largest singular value $(\sigma)$ for the matrix $\bfF$. Thus, given a video, RP signal is extracted through a singular value decomposition of $\bfF$. Finally, a narrow band-pass filter, centered around the possible range of respiration rate (5-30 bpm) is used to filter out other noise in the video, to get a robust estimate of the RP.

\subsection{VAD from Respiration Pattern}
\begin{figure}[t]
\centerline{\includegraphics[keepaspectratio, width=\columnwidth]{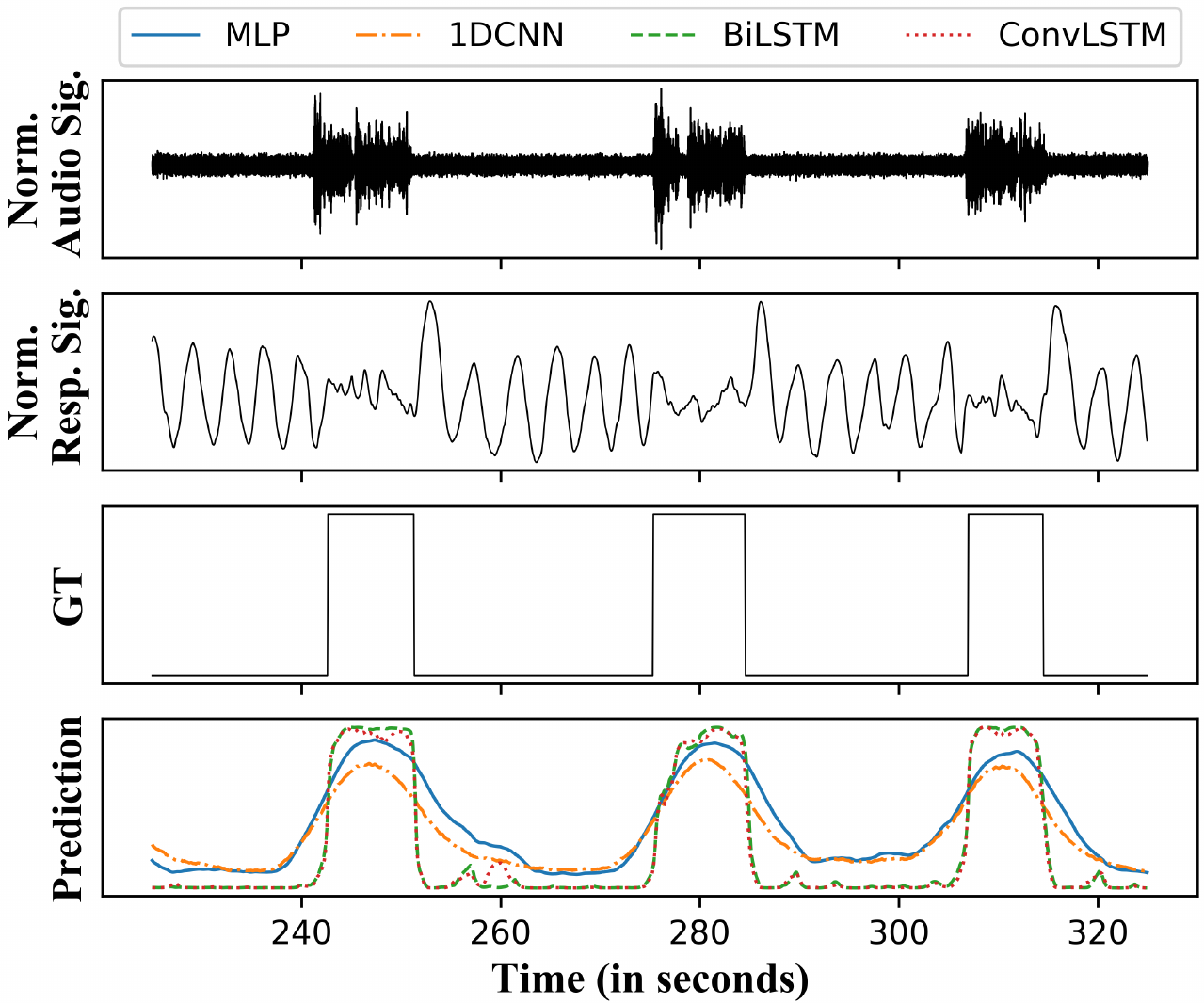}}
\caption{A segment of audio signal with the corresponding respiratory pattern, ground truth and predictions of all four models. While all models outputs a higher value of predictions during speech episodes, LSTM based models predicts the transitions much accurately.}
\label{aud-resp-gt-pred}
\end{figure}

After the extraction of the RP signal, the subsequent task is to perform VAD using it. Let $\bfR^*=[\bfr_1,\bfr_2,...,\bfr_N]$ denote the input RP sequence and $\mathbf{Y}^*=[\mathbf{y}_1,\mathbf{y}_2,...,\mathbf{y}_N]$ denote the binary valued output sequence such that, 1 where there is speech and 0 elsewhere. The objective of a VAD system is to predict $\mathbf{Y}^*$ given $\bfR^*$. This is cast as a sequence-to-sequence (Seq2Seq)  prediction problem with scalar-valued input and output sequences. Since the RP sequence is quasi-stationary, the sequences are chunked into smaller sub-sequences (with or without overlapping strides) and each of them is treated as an independent data point for processing. To solve the aforementioned Seq2Seq problem within a supervised learning setting, four neural discriminative models are proposed. Detailed description of the architectures can be found in the supplementary material.
\begin{enumerate}
    \item \textbf{MLP:} In the first model, a time-distributed Multi-Layer Perceptron (MLP) is proposed. It utilizes a feed forward Neural network with shared hidden layers across every temporal slice of the input. This is followed by a fully connected layer with sigmoid activation for prediction.
    \item \textbf{1DCNN:} In this model, the input signal is repeated and passed through time-distributed 1-D Convolutional layers  \cite{oord2016wavenet} followed by time-distributed dense layers for the final prediction.
    \item \textbf{BiLSTM:} Here, a bi-directional LSTM \cite{graves2005framewise} layer accepts the input signal and passes it to a time-distributed dense layer to preserve the one-to-one relationship between the input and the output. The final layer is again a time-distributed dense layer consisting of a single neuron with sigmoid activation for prediction of the speech activity. 
    \item \textbf{ConvLSTM:} In the final model, the input signal is passed through 1-D Convolutional layers, and the features thus extracted are fed to bi-directional LSTM layers. The final and the penultimate layer are time-distributed dense layers.
\end{enumerate}{}
Fig. \ref{fig:bd} gives the overview of the end-to-end workflow of the proposed approach. 
The first model is a naive MLP while the rest have been successfully used in a variety of time series modelling tasks. Data is created from a given RP series by using two types of chunking procedure as described next. Non-overlapping (NO): In this case, the RP series is chunked into non-overlapping data sub-sequences. During inference, the output of each of the sub-sequence is simply concatenated. Overlapping (O): Here, data is created with an overlapping windowing method with stride size of one. The output probabilities are averaged over all overlapping samples during inference. Further, since there is severe class imbalance, weighted binary cross entropy is used as the loss function for training all models. Fig. \ref{aud-resp-gt-pred} depicts a sample segment of speech with the corresponding RP signal, the ground truth and the predictions of all four models. It is seen that model with LSTMs predicts the transitions much sharper as compared to the other two models. Since the RP is used to perform a VAD task in the proposed method, it is termed as RespVAD.

% \begin{figure}[t]
% \centerline{\includegraphics[trim=25 1 40 34,clip,width=\columnwidth, keepaspectratio]{noisy_vid_pred.png}}
% \caption{Normalized respiratory pattern extracted from a noisy video is plotted with normalized audio signal. RespVAD predicts using the respiratory pattern and another visual VAD method KB \cite{dov2016kernel} predicts using the audio signal and video of the lips region.}
% \label{noisy-vid}
% \end{figure}

\begin{figure}[t]
\centerline{\includegraphics[trim=55 7 60 40,clip,keepaspectratio,width=\columnwidth]{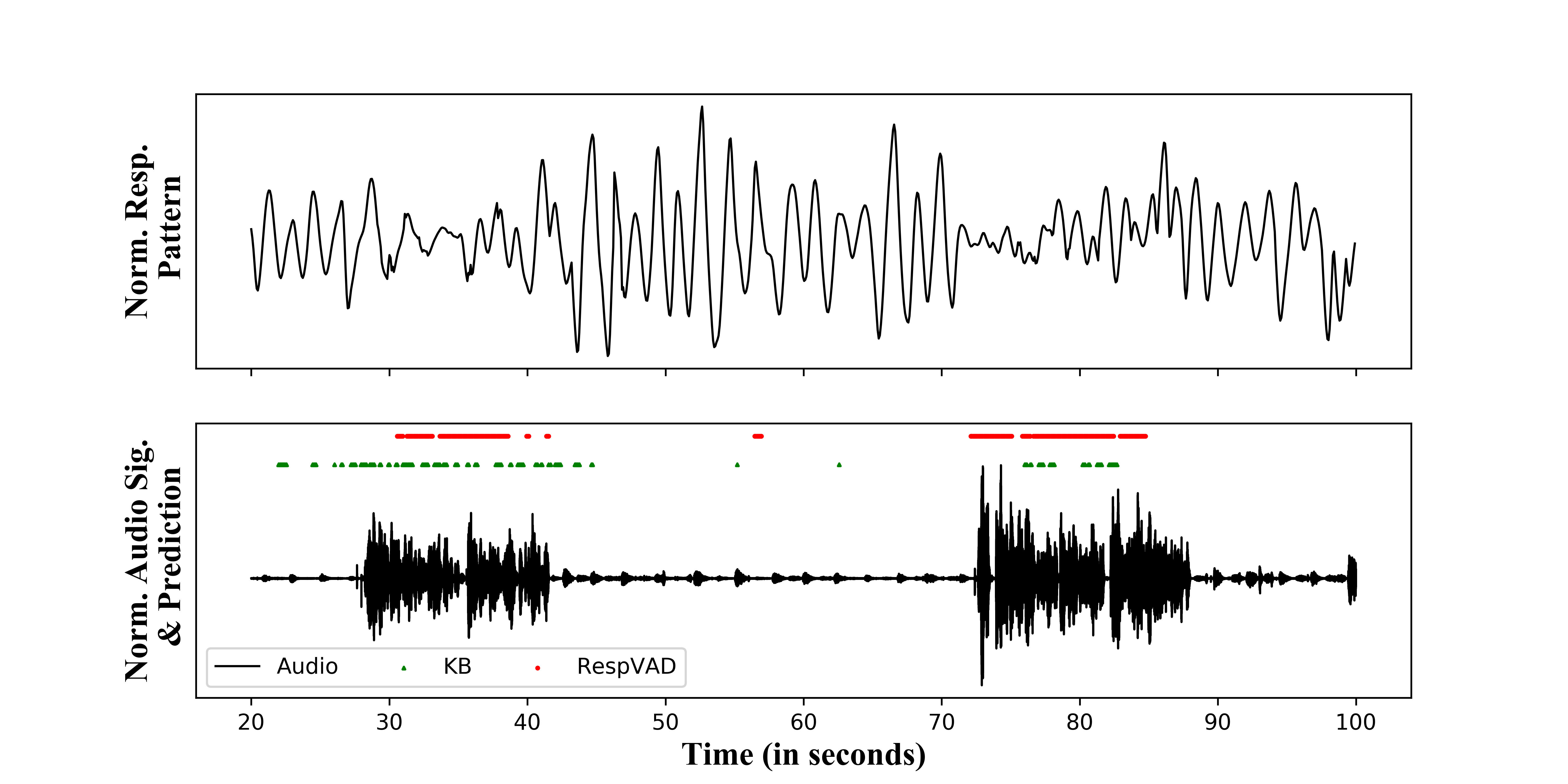}}
\caption{Normalized respiratory pattern extracted from a noisy video (subject is constantly moving) is plotted with normalized audio signal. VAD predictions by RespVAD and KB \cite{dov2016kernel} are marked as well.}
\label{noisy-vid-2}
\end{figure}

\begin{table*}[ht]
 \caption{Comparison of the performance of the models in RespVAD. It is seen that the ConvLSTM model with data overlap performs the best.}
    \begin{center}
	{
       {%
        \begin{tabular}{l c c c c c}
            \hline
            & Accuracy & Precision & Recall & F1 & AuROC \\
            \hline\hline
MLP (NO) & $0.796 \pm 0.087$ &  $0.608 \pm 0.131$ &  $0.821 \pm 0.04$ &   $0.688 \pm 0.086$ &  $0.88 \pm 0.036$ \\
MLP (O) & $0.85 \pm 0.093$ &  $0.682 \pm 0.145$ &  $0.939 \pm 0.023$ &   $0.779 \pm 0.103$ &  $0.944 \pm 0.031$ \\
1DCNN (NO) & $0.806 \pm 0.053$ &  $0.607 \pm 0.095$ &  $0.799 \pm 0.043$ &   $0.685 \pm 0.069$ &  $0.873 \pm 0.029$ \\
1DCNN (O) & $0.906 \pm 0.061$ &  $0.775 \pm 0.132$ &  $0.952 \pm 0.028$ &   $0.847 \pm 0.083$ &  $0.973 \pm 0.023$ \\
BiLSTM (NO) & $0.882 \pm 0.054$ &  $0.752 \pm 0.126$ &  $0.873 \pm 0.044$ &   $0.799 \pm 0.07$ &  $0.951 \pm 0.02$ \\
BiLSTM (O) & $0.926 \pm 0.049$ &  $0.824 \pm 0.122$ &  $0.946 \pm 0.021$ &   $0.875 \pm 0.073$ &  $0.981 \pm 0.011$ \\
ConvLSTM (NO) & $0.9 \pm 0.043$ &  $0.78 \pm 0.11$ &  $0.891 \pm 0.031$ &   $0.827 \pm 0.064$ &  $0.959 \pm 0.02$ \\
ConvLSTM (O) & $\boldsymbol{0.933 \pm 0.041}$ &  $\boldsymbol{0.835 \pm 0.108}$ &  $\boldsymbol{0.948 \pm 0.02}$ &   $\boldsymbol{0.884 \pm 0.064}$ &  $\boldsymbol{0.983 \pm 0.01}$ \\
            \hline

        \end{tabular}
        }}
    \end{center}
    \label{table:respvad_comparison_table}
\end{table*}

\begin{table}[ht]
 \caption{Comparison of the performance of RespVAD with four state-of-the-art algorithms for VAD. The first two are Audio-only based and the latter are Visual VAD methods.}
    \begin{center}
	{
        \begin{tabular}{l c c c c c}
            \hline
            Metric & SF \cite{drugman2015voice} & ED \cite{kim2018voice} & KB \cite{dov2016kernel} & DM \cite{dov2015audio}  &  RespVAD \\
            \hline \hline	
            Accuracy    & $0.871$   & $0.916$              & ${0.914}$  &  $0.799$   &  $\boldsymbol{0.933}$\\
            Precision   & $0.721$   & $\boldsymbol{0.862}$ & $0.825$    &  $0.711$   &  $0.835$\\
            Recall      & $0.928$   & $0.814$              & ${0.874}$  &  $0.745$   &  $\boldsymbol{0.948}$\\
            F1          & $0.809$   & $0.832$              & $0.841$    &  $0.718$   &  $\boldsymbol{0.884}$\\
            AuROC       & $0.945$   & $0.962$              & $0.966$    &  $0.952$   & $\boldsymbol{0.983}$\\
            \hline

        \end{tabular}
        }
    \end{center}
    \vspace{-4mm}
    \label{table:comparison_table}
\end{table}

\section{Experiments and Results}
\label{sec:expt}
\subsection{The Dataset}
A dataset consisting of video recordings from multiple devices such as laptop and cell-phone cameras of 50 human speakers was collected with consents. Each speaker was asked to speak non-uniformly spaced random sentences, of durations ranging from 2-15 seconds. Videos were recorded in natural environments such as office spaces, dormitories, class rooms, cafeteria etc. along with several sources of natural audio noises. The camera was placed at distances of about 3-5 ft with a focus on the subjects' face-to-abdomen region. Each video is of duration of 4-7 minutes with about ten speech episodes of varying length, recorded at a frame rate of 30 fps. Figure \ref{fig:videoframe} in the supplementary material presents two random video frames of two speakers with their consent.

\subsection{Experimental Details}
The dataset is split into four different train and test sets such that every split has $12$ non-overlapping speakers. \footnote{The variation of the performance as a function of percentage of training data is shown in Fig. \ref{fig:ttp} (a) of the supplementary material.}. 
The rationale behind such  splits is to avoid any speaker specific bias in the data. A technique as outlined in Section II is used to generate respiration signals from the videos and create data chunks of a fixed width of 100 samples
\footnote{It was empirically observed that the performance saturates after a width of 100 samples as shown in Fig.\ref{fig:ttp} (b) in the Supplementary material.}. 
For each split, two types of datasets have been created with overlapping and non-overlapping chunks. Figure \ref{fig:o_no_dataset} in the supplementary material provides an illustration of the two types of datasets. Since there is a considerable class imbalance, accuracy alone doesn't fully quantify the performance. Consequently, Precision, Recall, F1 score and AuROC along with accuracy is considered for evaluation. The average and standard deviation values of all the metrics over ten runs is reported. For accuracy computation, the raw predictions are thresholded at $0.5$. 

\begin{figure}[t]
    \centerline{\includegraphics[width=\columnwidth,keepaspectratio]{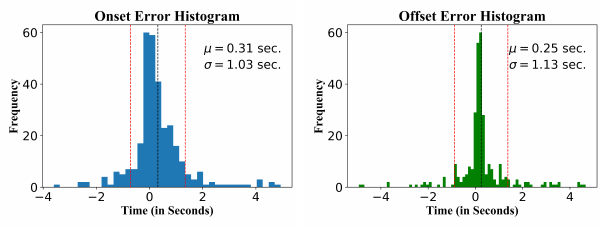}}
    \caption{Histogram for the error committed by RespVAD at non-speech to speech (onset) and speech to non-speech transitions (offset). In both cases, the peak is around 0.25-0.3s with a standard deviation $< 1.2$s. }
    \label{fig:err_hist}
\end{figure}

\subsection{Results and Comparison}

Table \ref{table:respvad_comparison_table} lists the performance of  four RespVAD models averaged over all the splits.  It is seen that the ConvLSTM model performs consistently the best amongst all models followed by BiLSTM, 1DCNN and MLP.  This suggests that the Convolutional block extracts useful features that are beneficial for prediction compared to an LSTM with raw speech as the input as in \cite{oord2016wavenet}. It is also observed that the models with LSTM blocks better than the ones without it confirming the effectiveness of Recurrent models in modeling time series over MLPs. Further, every model performs better on the overlapping dataset for a given split because of the ensemble effect. \\

To further quantify the resolution of the detections, the histogram of the errors at every transition are plotted. That is, for every speech-to-non-speech (Offset) or non-speech-to-speech (Onset) transitions in the ground truth, the time difference between the GT and the predictions are measured and the histograms of all such differences are plotted in  Fig. \ref{fig:err_hist}. It is seen that in both the cases of the onset and the offset, the mean difference is around 0 with a standard deviation of 1 second. This implies that the RespVAD detects the speech within a range of 1 second before/after its actual occurrence. \\
The performance of RespVAD is compared against four state-of-the-art VADs algorithms namely: (a) Source+Filter (SF) model \cite{drugman2015voice}, (b) Encoder-Decoder (ED) model \cite{kim2018voice}, (c) Kernel-based (KB) model \cite{dov2016kernel} and (d) Diffusion maps (DM) model \cite{dov2015audio}. The former two are audio-only methods while the latter two are Visual VADs. For all the baselines, the original implementation provided by the Authors are used. Table \ref{table:comparison_table} lists the average performance metrics for all the algorithms. It is observed that despite not using the audio data, RespVAD performs the best under majority of the metrics. The classical signal processing based method (SF) has the least performance since the audio data is recorded in an environment with natural non-stationary noise at around 10-20 dB SNR. Further, RespVAD has an advantage over other Visual VAD techniques as there is no need to detect a specific region of interest (such as lips/mouth) since the RP is shown to be able to be extracted in oblique angles as well \cite{prathosh2017estimation,AC}. These result demonstrate the effectiveness of RespVAD for audio-independent VAD task.

To demonstrate the feasibility of RespVAD on videos with high levels of video noise, in Fig. \ref{noisy-vid-2}, the RP with the VAD predictions are shown on a video where the subject is randomly moving. It is seen that RespVAD performs reasonably well despite high amounts of video noise. This is because of the robustness of the RP extraction algorithm. Nevertheless, extraction of RP with very high levels of video noise is a subject of future work.

\section{Conclusion}

In this paper, a new paradigm for Voice Activity Detection based on the changes that occur in the respiration pattern during speech episodes is proposed. Respiration pattern provides additional evidence for VAD especially in audio-noise-heavy environments. Future directions may aim at (i) combining all three modalities (of audio, lip movement and respiration) for improving VAD and (ii) extracting the linguistic content of speech from the RP as hypothesized in \cite{ladefoged1968linguistic}.

% put at least one blank line to end the scriptsize paragraph and
% then revert back to normalsize.
\normalsize

% Last page column equalization
%
% IEEE Sensors Letters does balance the columns on the last page.
% Can use:
% \IEEEtriggeratref{8}
% to trigger a \newpage just before the given reference number to
% balance the columns on the last page. Adjust the reference number
% as needed - this may need to be readjusted if the document is 
% modified later.
% Alternatively, you can also directly use something like
% \enlargethispage{-7in}
% on the last page instead of breaking at a specific reference number.

%\bibliographystyle{IEEEtran}
% Before submitting to IEEE Sensors Letters, manually copy in the
% resultant .bbl file contents in place of the \bibliographystyle and
% \bibliography lines here:

\bibliographystyle{ieeetr}
\bibliography{references}

\section{Supplementary Material}
\subsection{Derivation of Equation \ref{eqn:optimalR}}
The objective function as presented in Section \ref{sec:proposed_method}, Equation \ref{eqn:objectivefn} is a constrained optimization problem. 
\begin{equation}
\bfR^* = \underset{{\bfR:\ ||\bfR||=1}}{\arg\max} \ \   ||\bfF\bfR||^2
\label{eqn:supp_objectivefn}
\end{equation}
The lagrangian of Equation \ref{eqn:supp_objectivefn} can be written as:
\begin{equation}
    f: \bfR^T\bfF^T\bfF\bfR + \lambda(\bfR^T\bfR - 1)
    \label{eqn:lagrangian}
\end{equation}
Differentiating Equation \ref{eqn:lagrangian} w.r.t. $R$ and equating with zero,
\begin{equation}
    \bfF^T\bfF\bfR + \lambda \bfR = 0
    \label{eqn:eigenrelation}
\end{equation}{}
For Equation \ref{eqn:eigenrelation} to be satisfied $\bfR$ must be an eigenvector of $\bfF^T\bfF$.
Now the expression $\bfR^T\bfF^T\bfF\bfR$ can be written as:
\begin{equation}
    \begin{split}
        \bfR^T\bfF^T\bfF\bfR &= \sigma \bfR^T\bfR \\
        &= \sigma ~~ [\because ||\bfR|| = 1]
    \end{split}{}
\end{equation}
Therefore, $\bfR^T\bfF^T\bfF\bfR$ is maximum when $\sigma$ is the largest eigenvalue of $\bfF^T\bfF$. Now let us consider the singular value decomposition of $\bfF$.
\begin{equation}
    \begin{split}
        \bfF &= \mathbf{U}\Sigma \mathbf{V}^T \\
        \mathbf{U}^T\bfF &= \Sigma \mathbf{V}^T
    \end{split}{}
\end{equation}{}
The columns of $\mathbf{V}$ (right-singular vectors) i.e. the rows of $\mathbf{V}^T$ are the Eigenvectors $\bfF^T\bfF$. Let, $\sigma$ denote the largest eigenvalue of $\bfF^T\bfF$ and $\bfR^*$ denotes the corresponding Eigenvector. Let, $\mathbf{U}_1$ denote the corresponding Eigenvector of $\bfF\bfF^T$
Therefore, 
\begin{equation}
    \bfR^* =\frac{1}{\sigma}\mathbf{U}_1^T\bfF
\end{equation}.

\subsection{Sample Video Frames from the Dataset}
\begin{figure}[!htbp]
\centering
\begin{subfigure}{\columnwidth}
  \centering
  \includegraphics[keepaspectratio, width=0.9\columnwidth]{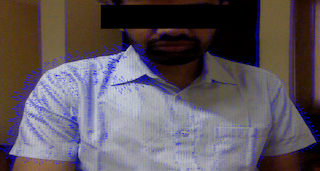}
  \caption{A frame from the video of a male speaker.}
  \label{fig:male_speaker}
\end{subfigure}

\begin{subfigure}{\columnwidth}
  \centering
  \includegraphics[keepaspectratio, width=0.9\columnwidth]{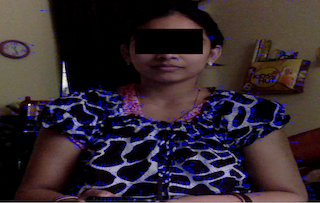}
  \caption{A frame from the video of a female speaker.}
  \label{fig:female_speaker}
\end{subfigure}
\caption{Depiction of sample frames from the dataset with the corresponding flow vectors (in blue). It is seen that the majority of the flow components emerge from regions where motion caused from respiration are predominant.}
\label{fig:videoframe}
\end{figure}

In this section, we present two sample frames from the video dataset along with the flow vectors. Figure \ref{fig:male_speaker} presents video frame of a male speaker and Figure \ref{fig:female_speaker} presents video frame of one female speaker.

\subsection{Overlapped and Non-overlapped Dataset}
\begin{figure}[!htbp]
\centerline{\includegraphics[keepaspectratio, width=\columnwidth]{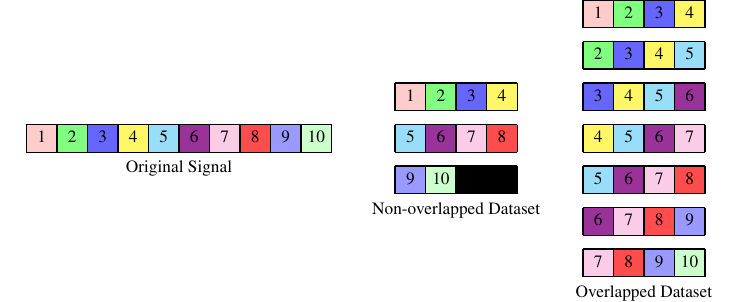}}
\caption{Illustration of non-overlapped and overlapped dataset creation. In this example, the original signal has a length $10$ and window size is chosen to be $4$. The last sample in the non-overlapped dataset uses zero-padding to maintain the chunk size.}
\label{fig:o_no_dataset}
\end{figure}

As described in the main paper, we have experimented using one overlapped dataset and one non-overlapped dataset. Figure \ref{fig:o_no_dataset} presents an illustration of how these two types of datasets are created.

\subsection{Variation of Performance w.r.t. Sample Complexity and Chunk Size}
\begin{figure}[!htbp]
  \centerline{\includegraphics[keepaspectratio, width=\columnwidth]{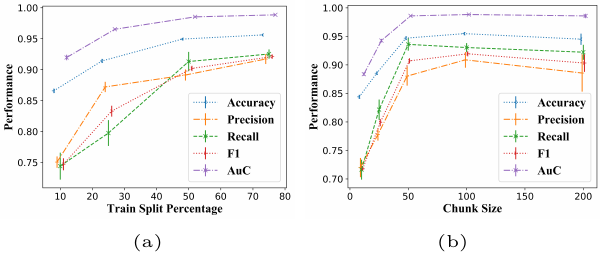}}
   \caption{Variation of performance metrics of the ConvLSTM model with respect to the train-test split ratio and chunk size for the overlapping case.}
  \label{fig:ttp}
\end{figure}
Figure \ref{fig:ttp} presents the effect of training split size and chunk size on the performance of the proposed ConvLSTM model.

\subsection{ROC and Precision-Recall Plots}
\begin{figure}[!ht]
\centerline{\includegraphics[keepaspectratio, width=\columnwidth]{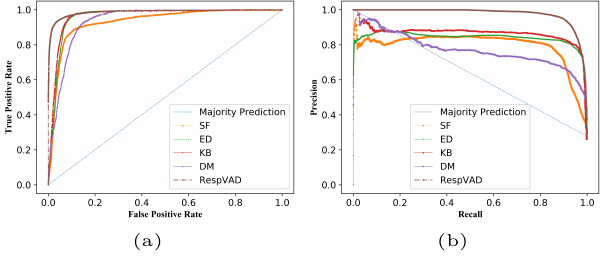}}
\caption{ROC and Precision-Recall Curves for RespVAD and other methods using the test set of a single split. It is seen that RespVAD consistently performs better.}
\label{fig:ROC_PR}
\end{figure}

It is seen in Figure \ref{fig:ROC_PR}, the proposed method RespVAD outperforms other baseline methods as reflected by ROC Curve and Precision/Recall plots. The plot is based on the performance of the models on the test set of a single split.

\subsection{Model Architectures}
\subsubsection{MLP}
\begin{equation*}
    \begin{split}
        & \boldsymbol{R^*}\in\mathbb{R}^{w}, \boldsymbol{Y^*}\in\{0, 1\}^{w}\\
        &\to \text{RepeatVector}(100) \\
        &\to \text{TimeDist}(\text{FC}_{128}) \to \text{ReLU} \\
        &\to \text{TimeDist}(\text{FC}_{64}) \to \text{ReLU} \\
        &\to \text{TimeDist}(\text{FC}_{64}) \to \text{ReLU} \\
        &\to \text{TimeDist}(\text{FC}_{64}) \to \text{ReLU} \\
        &\to \text{TimeDist}(\text{FC}_{1}) \to \text{Sigmoid}
    \end{split}
\end{equation*}

\subsubsection{1DCNN}
\begin{equation*}
    \begin{split}
        & \boldsymbol{R^*}\in\mathbb{R}^{w}, \boldsymbol{Y^*}\in\{0, 1\}^{w}\\
        &\to \text{RepeatVector}(30) \\
        &\to \text{Reshape}(w, 30, 1)\\
        &\to \text{TimeDist}(\text{Conv1D}_{32, 3, 1}) \to \text{tanh} \\
        &\to \text{TimeDist}(\text{Conv1D}_{32, 3, 1}) \to \text{tanh} \\
        &\to \text{Flatten} \\
        &\to \text{TimeDist}(\text{FC}_{64}) \to \text{ReLU} \\
        &\to \text{TimeDist}(\text{FC}_{128}) \to \text{ReLU} \\
        &\to \text{TimeDist}(\text{FC}_{1}) \to \text{Sigmoid}
    \end{split}
\end{equation*}
Here, $\text{Conv1D}_{32, 3, 3}$ implies that the layer consists of $32$ filters of kernel size $3$ and dialation rate $1$.

\subsubsection{BiLSTM}
\begin{equation*}
    \begin{split}
        & \boldsymbol{R^*}\in\mathbb{R}^{w\times 1}, \boldsymbol{Y^*}\in\{0, 1\}^{w\times 1}\\
        &\to \text{BiLSTM}_{128} \\
        &\to \text{BiLSTM}_{128} \\
        &\to \text{TimeDist}(\text{FC}_{32}) \to \text{ReLU} \\
        &\to \text{TimeDist}(\text{FC}_{1}) \to \text{Sigmoid}
    \end{split}
\end{equation*}

\subsubsection{ConvLSTM}
\begin{equation*}
    \begin{split}
        & \boldsymbol{R^*}\in\mathbb{R}^{w\times1\times1}, \boldsymbol{Y^*}\in\{0, 1\}^{w\times1}\\
        &\to \text{TimeDist}(\text{Conv1D}_{16, 5, 3}) \to \text{tanh} \\
        &\to \text{TimeDist}(\text{Conv1D}_{16, 5, 3}) \to \text{tanh} \\
        &\to \text{Flatten} \\
        &\to \text{BiLSTM}_{128} \\
        &\to \text{BiLSTM}_{128} \\
        &\to \text{TimeDist}(\text{FC}_{32}) \to \text{ReLU} \\
        &\to \text{TimeDist}(\text{FC}_{1}) \to \text{Sigmoid}
    \end{split}
\end{equation*}

$w$ denotes the window size. $w=100$ in all the experiments.

\end{document}